\documentclass[conference]{IEEEtran}
\IEEEoverridecommandlockouts
\usepackage{cite}
\usepackage{amsmath,amssymb,amsfonts}
\usepackage{algorithmic}
\usepackage{graphicx}
\usepackage{textcomp}
\usepackage{xcolor}
\def\BibTeX{{\rm B\kern-.05em{\sc i\kern-.025em b}\kern-.08em
    T\kern-.1667em\lower.7ex\hbox{E}\kern-.125emX}}
\begin{document}

\title{Lightweight Gaze Estimation Model Via Fusion Global Information
}

\author{\IEEEauthorblockN{Zhang Cheng}
\IEEEauthorblockA{\textit{College of Computer and Information Science} \\
\textit{Chongqing Normal University}\\
Chongqing, China\\
czhang2026@163.com}
\and
\IEEEauthorblockN{Yanxia Wang\textsuperscript{*}}
\IEEEauthorblockA{\textit{College of Computer and Information Science} \\
\textit{Chongqing Normal University}\\
Chongqing, China\\
wangyanxia@cqnu.edu.cn;}

}

\maketitle

\begin{abstract}
Deep learning-based appearance gaze estimation methods are gaining popularity due to their high accuracy and fewer constraints from the environment. However, existing high-precision models often rely on deeper networks, leading to problems such as large parameters, long training time, and slow convergence. In terms of this issue, this paper proposes a novel lightweight gaze estimation model FGI-Net(Fusion Global Information). The model fuses global information into the CNN, effectively compensating for the need of multi-layer convolution and pooling to indirectly capture global information, while reducing the complexity of the model, improving the model accuracy and convergence speed. To validate the performance of the model, a large number of experiments are conducted, comparing accuracy with existing classical models and lightweight models, comparing convergence speed with models of different architectures, and conducting ablation experiments. Experimental results show that compared with GazeCaps, the latest gaze estimation model, FGI-Net achieves a smaller angle error with 87.1\%  and 79.1\% reduction in parameters and FLOPs, respectively (MPIIFaceGaze is 3.74°, EyeDiap is 5.15°, Gaze360 is 10.50° and RT-Gene is 6.02°). Moreover, compared with different architectural models such as CNN and Transformer, FGI-Net is able to quickly converge to a higher accuracy range with fewer iterations of training, when achieving optimal accuracy on the Gaze360 and EyeDiap datasets, the FGI-Net model has 25\% and 37.5\% fewer iterations of training compared to GazeTR, respectively.Our code is published at \textcolor{blue}{ https://github.com/CZ178/FGI-Net}
\end{abstract}

\begin{IEEEkeywords}
gaze estimation, lightweight, fusion global information, CNN, Transformer 
\end{IEEEkeywords}
\section{Introduction}Gaze is an important nonverbal communication method that plays a crucial role in social interaction, emotional expression, and information transmission. For example, in education by analyzing students' eye movement data, it is possible to understand the distribution of their attention and points of interest during the learning process, thus realizing personalized teaching. In medicine, eye movement analysis can be used to pre-diagnose neurological disorders such as Autism Spectrum Disorder(ASD)\cite{b1}. As a result, gaze estimation has gradually become a research topic. With the development of deep learning, there is an increasing number of deep learning-based models for gaze estimation. However, existing gaze estimation models have problems such as large model size, long training time, and slow convergence, which limits the practical application of this technology. Therefore, the study of lightweight gaze estimation models with fast convergence and high accuracy is of particular importance.

Currently, deep learning-based gaze estimation models mostly use CNN or Transformer architectures. For example, Cai et al. \cite{b3} propose a model called UnReGA, which is suitable for across-domain gaze estimation tasks, using ResNet18 as the backbone network. Balim et al. \cite{b5} propose an end-to-end Frame-to-Gaze model, which can be well generalized to extreme camera view changes. However, the structure is complex and the camera-to-screen geometry must be known a priori, which imposes certain limitations that share with data normalization-based works. O'Shea et al. \cite{b7} propose a model called SwinIR that combines ResNet18 and Swin Transformer to improve the quality of eye images using super-resolution technology to enhance the performance and accuracy of gaze estimation systems. But SwinIR requires significant computational resources and time, which makes its practical application scenarios limited, and factors such as illumination variations and environmental noise in real-world can affect the robustness of super-resolution techniques. To alleviate some of these problems, lightweight gaze estimation models have become a popular research topic.

To make them suitable for a wider range of application scenarios without consuming more resources, the paper proposes a lightweight gaze estimation model FGI-Net that \textbf{F}uses \textbf{G}lobal \textbf{I}nformation. The FGI-Net model expands the receptive field to fuses global information so that it captures a wider range of context information to improve the accuracy of the model.  Meanwhile, the fusion of global information effectively replaces the way of expanding the receptive field by stacking the convolutional layer and pooling layer to reduce of convolutional layer and pooling layer significantly, therefore, the parameters and FLOPs of the model reduce to make the model converge quickly with less iterative training.

The main contributions of this paper are as follows:
\begin{itemize}
\item A Global Information Fusion module is proposed. The module not only has a powerful feature extraction capability but also focuses more on useful channel information by learning adaptive channel weights. In addition, the shifting window mechanism allows the module to effectively implement global information interaction, which fully utilizes the features related to gaze estimation, thus improving the model accuracy.
\item A large number of experiments have been conducted to verify the performance of the FGI-Net model. This model achieves lightweight while improving accuracy, making it more advantageous to deploy in resource-limited environments, especially in mobile devices and embedded systems.
\end{itemize}

\section{Related Work}
\subsection{Gaze Estimation}
Gaze estimation methods are mainly divided into two categories: model-based and appearance-based. Model-based methods rely on the physiological structure and motion characteristics of the eyes for gaze estimation, such as corneal reflex\cite{b8}\cite{b9}, pupil position\cite{b10}, etc. This approach is more accurate, but requires accurate parameters and model assumptions, as well as personalized adjustments tailored to different individual needs, thereby increasing the complexity and computational cost of the model. It performs well under controlled conditions, but the performance of the model deteriorates in complex and constantly changing environments. In addition, specialized equipment is also required, such as infrared cameras. The appearance-based method utilizes facial images \cite{b11}\cite{b12} or eye images \cite{b13}\cite{b14} for gaze estimation, directly extracting features from the input image, and learning the mapping function from appearance to gaze direction. It does not require specialized equipment and is more widely used.

At present, appearance-based gaze estimation methods have received much attention. Zhang et al. \cite{b15} use a 5-layer shallow convolutional neural network LeNet to estimate the gaze direction. Since the LeNet network is too simple, it makes the model lacking in goodness-of-fit and feature extraction capabilities. Therefore, Cheng et al. \cite{b17} use ResNet18 as the backbone network of the model, significantly improving the accuracy of the model. To further improve the performance of the model and make it more robust, Abdelrahman et al. \cite{b18} use ResNet50 to predict the gaze direction and regress each gaze angle separately to improve the prediction accuracy of each angle and thus improve overall performance. However, as the performance of the model improves, the complexity also increases, requiring more computing resources. Therefore, the current focus is on optimizing performance while lightweight the model to adapt to resource-limited scenarios.

\subsection{Lightweight network}
Existing methods for gaze estimation have a number of parameters and high computational complexity, which have some limitations for practical applications. Therefore, lightweight models are receiving increasing attention. The proposal of parameter reduction techniques such as dilated convolutional \cite{b20} and depthwise separable convolution\cite{b27}, as well as lightweight models such as EfficientNet \cite{b25}, SqueezeNet\cite{b4}, ShuffleNet\cite{b16}, and MobileVit\cite{b22}, has to some extent alleviated these problems. For example, Chen et al. \cite{b19} use Dilated Convolutional instead of ordinary convolution, which significantly reduces the parameters of the model. However, a high dilation rate can lead to information loss and increase the dependence of network decisions on distant pixels, thereby reducing the interpretability of the model. Especially when using only eye region images as input, a high dilation ratio may overlook important information, which may lead to significant performance degradation for regression problems such as gaze estimation. Xu et al. \cite{b21} propose a lightweight network FR-Net for gaze estimation, which uses MobileVit-v3 as the backbone network and introduces Fast Fourier Transform (FFT) Residual Block to extract relevant frequency and spatial features related to the eyes, effectively solving the problem of information loss. However, due to the integration of CNN, Transformer, and FFT architectures in FR Net, the model is susceptible to hardware and matching operators, resulting in issues such as long inference time. On the other hand, existing lightweight models either only integrate global information from high-level features that have been downsampled several times, or indirectly obtain global information through stacked convolutional and pooling layers. However, the high-level features after downsampling are not sensitive to subtle changes in the input image, and the pooling layer may also lose information, which may have a significant impact on the regression task.

\begin{figure*}[htbp]
\centerline{\includegraphics[width=0.9\linewidth]{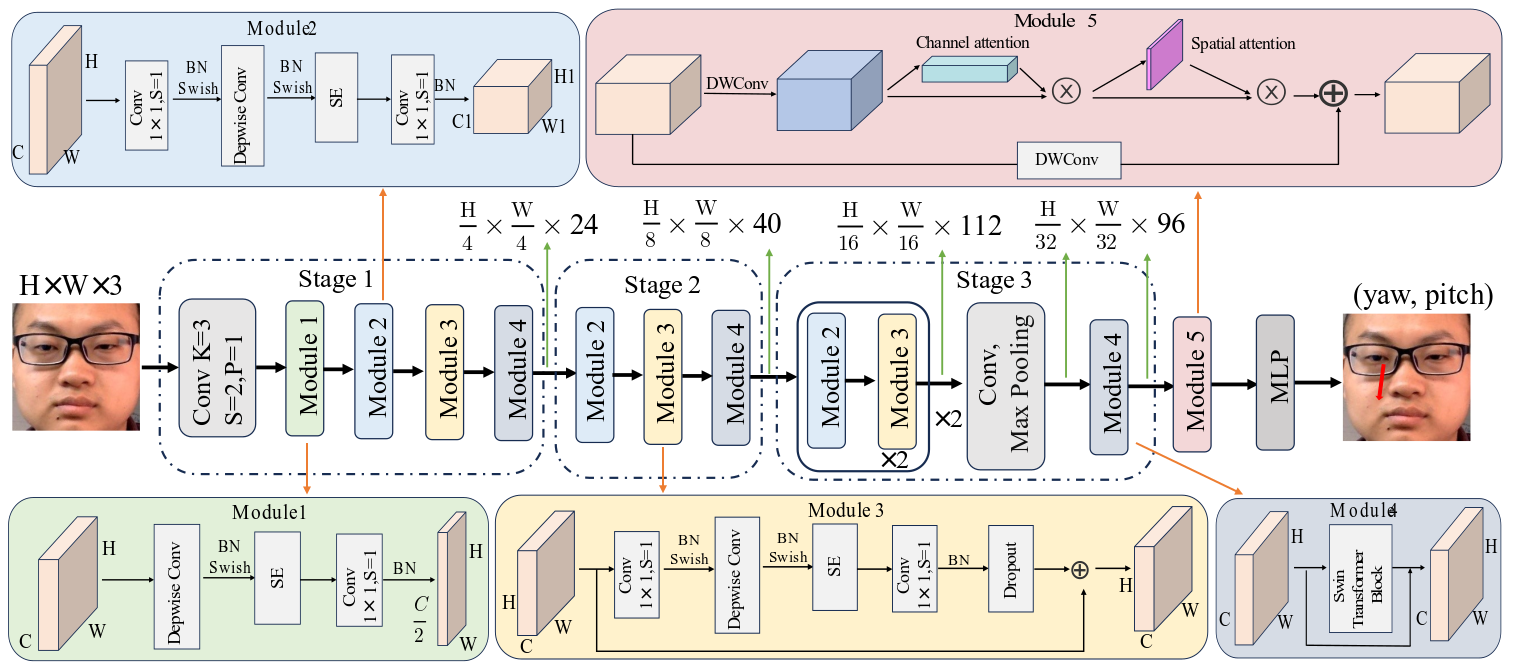}}
\caption{FGI-Net network structure.}
\label{fig1}
\end{figure*}
\section{Methods}
The goal of appearance-based gaze estimation is to predict the gaze direction using the input image. This section details the FGI-Net network model proposed in this paper, which fuses the powerful feature extraction capability of CNN and the advantages of Swin Transformer's larger receptive field by alternating between CNN and Swin Transformer. The FGI-Net model can be categorized into the Global Information Fusion module (Stage 1, Stage 2, and Stage 3), Res\_CBAM (Module 5), and MLP, and its structure is shown in Fig.1.

The Global Information Fusion module proposed in this paper improves the shift window mechanism, expands the receptive field of the model, enhances the modeling ability of long-distance dependencies, and enables FGI-Net to be effectively applied to tasks that require understanding of global contextual information. To avoid overfitting of low-level features and loss of high-level semantics, linear decay dropout is used in three Global Information Fusion modules. In this module, CNN is used to extract local features, and global contextual information is fused through a shift window mechanism to provide prior information for gaze prediction. It can also be used to compensate for the gaze direction of the eyes, thereby improving the accuracy and robustness of the model. The Res\_CBAM module is based on CBAM improvements, which improve model's perception and utilization of eye region features while making the model more sensitive to internal changes in images. MLP is used for the regression gaze direction, which can learn and fit complex mapping relationships, thereby more flexibly adapting to various complex data distributions and patterns, making MLP a powerful tool for solving regression tasks.
\subsection{Global Information Fusion Module}
Models based on CNN architectures can only obtain global information indirectly by stacking convolutional and pooling layers to obtain a larger receptive field. However, the stacking of convolutional layers leads to an increase in the number of model parameters\cite{b43}\cite{b44}, and pooling leads to an information loss, Vision Transformer(ViT)\cite{b23} proposed by Dosovitskiy et al. effectively solves this problem. ViT leverages the self-attention mechanism of Transformer to achieve global perception, which is not subject to a fixed size of the local receptive field and is able to simultaneously take into account the global information of the input image and better understand the overall structure and context of the image. To further reduce the number of parameters and computational resources, Swin Transformer \cite{b6} proposed by Liu et al. reduces the computational complexity and saves a lot of computational overhead by introducing the shift window mechanism and downsampling mechanism. However, since the Transformer structure lacks the inductive bias of CNNs and requires large-scale datasets to exhibit better performance than CNNs, it becomes crucial to combine the respective advantages of CNNs and Transformers.

As we all know, to improve the accuracy of the model, we can only start from three dimensions: depth, width, and input image resolution. However, changing a single dimension often leads to performance saturation, Tan et al. propose EfficientNet \cite{b25} model for classification tasks, which adopts a composite scaling method and scales in three dimensions of network width, depth, and resolution at the same time, significantly improving the performance of the model, as shown in Fig.2. In order to preserve the advantages of EfficientNet and apply it to gaze estimation tasks, the FGI-Net network architecture(as shown in Fig.1) proposed in this paper is designed according to EfficientNet. Among them, Stage 1, Stage 2, and Stage 3 are three Global Information Fusion modules with different structures, whose main architectures are CNN and Swin Transformer. Modules 1, 2, and 3 have the same structure as the corresponding modules in EfficientNet, and module 4 is added on top of this. Module 4 is designed based on the shift window attention mechanism, which effectively extends the receptive field of the model and realizes global information interaction instead of stacking convolution and pooling to capture global information, thus achieving lightweight. After being processed by Module 3 in Stage 3, the input data has been downsampled 16 times, and the number of channels is 112. In order not to significantly increase parameters and FLOPs, dimension reduction and downsampling of the data are performed before the last global information interaction, and the feature map with the number of channels is 96 after 32 times of downsampling. Finally, in order to improve the nonlinear fitting ability of the model, the fully connected layer in FGI-Net is increased to two layers, and the number of hidden layer neurons is 32.

\begin{figure*}[htbp]
\centerline{\includegraphics[width=0.9\linewidth]{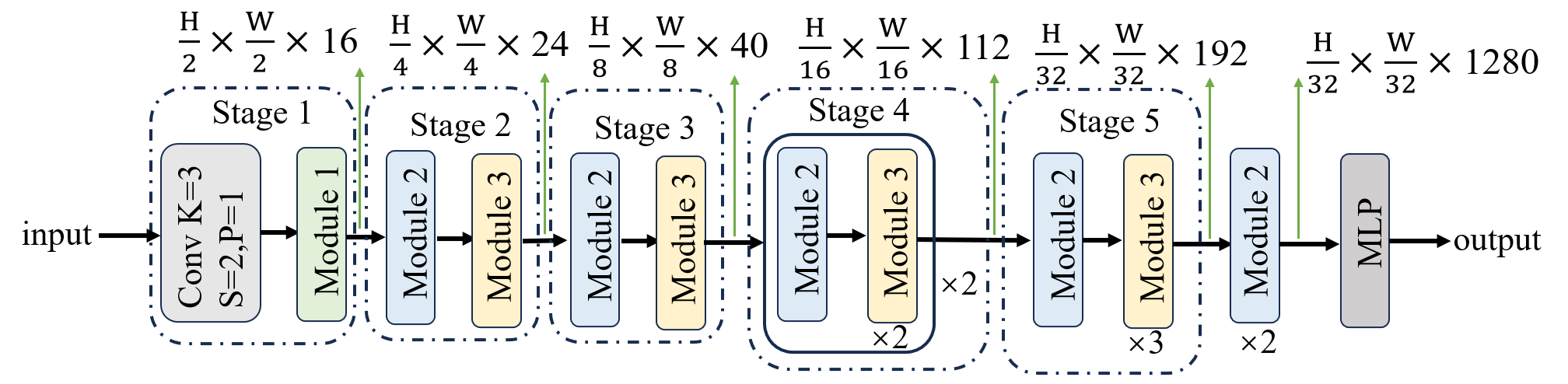}}
\caption{EfficientNet network structure.}
\label{fig2}
\end{figure*}

\begin{figure}[htbp]
\centerline{\includegraphics[width=0.8\linewidth]{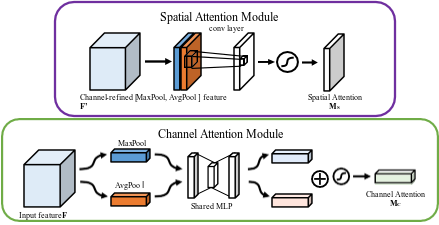}}
\caption{Top: Spatial Attention Module, Bottom: Channel Attention Module[26]}
\label{fig3}
\end{figure}
\subsection{ Res\_CBAM}
The Global Information Fusion module fuses feature information at the spatial level but ignores information exchange at the channel level. To address this issue, this paper proposes the Res\_CBAM (as shown in Fig.1 Module 5) module, which is improved based on CBAM\cite{b26} (as shown in Fig.3) and is divided into two parts: spatial attention and channel attention, while residual connections are added in order to accelerate the convergence of the model and prevent gradient disappearance.

The spatial attention module, which focuses the model on more important regions of the image and thus increases the model's attention to key information, is computed as follows:
\begin{equation}
M_s(F) = \sigma(f^{7\times7}([AvgPool(F);MaxPool(F)])) \label{eq1}
\end{equation}
where $F$ denotes the input features, $\sigma$ denotes the sigmoid function, $f^{7\times7}$ denotes the convolution kernel as a 7$\times$7 convolution operation, and $M_s(F)$ denotes the generated spatial attention map.

The channel attention module learns the correlation between different channels in an image to enable it to better understand the image content, calculated as follows:
\begin{equation}
M_c(F) = \sigma(MLP(AvgPool(F))+MLP(MaxPool(F))) \label{eq2}
\end{equation}
where $M_c(F)$ denotes the generated channel attention map.

The Res\_CBAM module is calculated as follows:
\begin{align}
 \begin{split}
F_{res} = f^{3\times3}_{DW}(F_{in})\\
F^{'} = f^{3\times3}_{DW}(F_{in})\\
F^{''} = M_c(F^{'})\otimes F^{'}\\
F^{'''} = M_s(F^{''})\otimes F^{''}\\
F_{out} = F_{res} + F^{'''}
  \end{split}
\end{align}
where $F_{in}$ and $F_{out}$ denote the input and output features, respectively, $f^{3\times3}_{DW}$ denotes the convolution kernel is a $3\times3$ depthwise separable convolution, and $\otimes$ denotes the element-by-element multiplication.

\section{Experimental setup and results}
In this section, angular error, Parameters, and FLOPs are used as evaluation metrics to experimentally validate FGI-Net based on widely used datasets. Among them, angle error refers to the angle difference between the predicted gaze direction and the actual gaze direction, usually measured in degrees(°). A smaller angle error indicates better model performance. This section mainly introduces the experimental setting, the comparison of accuracy between FGI-Net and recent methods, and the comparison of convergence speed with different architecture models. At the same time, ablation experiments are conducted to explore the impact of key components on the performance of FGI-Net.
\subsection{Experimental Settings}
This subsection focuses on the datasets used in the experiments and the implementation details.
\paragraph{Datasets}To train high-precision models and validate their performance, the researchers construct a series of gaze estimation datasets. This paper conducts experiments based on five datasets: ETH-XGaze(ETH) \cite{b28}, Gaze360(G) \cite{b29}, MPIIFaceGaze(M) \cite{b30}, EyeDiap(E)\cite{b31}, and RT-Gene(R) \cite{b32}, in which ETH-Xgaze is used as the pre-training dataset. The ETH-Xgaze dataset, acquired in a laboratory setting using a SLR camera, contains high-resolution images of different head poses of 110 subjects (47 female and 63 male) of different ethnicities and ages and covers 15 different illuminations. The Gaze360 dataset collects images of 238 subjects in five indoor scenarios (53 subjects) and two outdoor scenarios (185 subjects). The MPIIFaceGaze dataset is recorded over a period of several months in an everyday environment, covering complex and varied external environments, eye appearance, and illumination, and contains nearly 45k images from 15 subjects. The EyeDiap and RT-Gene datasets consist of 16k images from 14 subjects and 92K images from 15 subjects, respectively. Due to data occlusion and other conditions, the data used in this paper are pre-processed data, and thus the amount of data from the original text is somewhat different. The detailed data are shown in Table 1.

\begin{table}[htbp]
\caption{Gaze Estimation Dataset Details}
\begin{center}
\begin{tabular}{cccccc}
\hline
Datasets  & Subjects & Total & Head Pose & Gaze Angle & Time \\ 
    &  &   & (yaw, pitch)   & (yaw, pitch)  &               \\ \hline
ETH{\cite{b28}}    & 110               & 915K           & ±80°, ±80°             & ±120°, ±70°             & 2020          \\
G{\cite{b29}}      & 238               & 172K           & ±90°, N/A              & ±140°, -50°             & 2019          \\
M{\cite{b30}} & 15                & 45K            & ±15°, 30°              & ±20°, ±20°              & 2017          \\
E{\cite{b31}}      & 14                & 14K            & ±15°, 30°              & ±25°, 20°               & 2014          \\
R{\cite{b32}}      & 15                & 92K            & ±40°, ±40°             & ±40°, -40°              & 2018          \\ \hline
\end{tabular}
\label{tab1}
\end{center}
\end{table}

\paragraph{Implementation Details}The experimental input image size is 224×224×3, and the output result is a 3D vector, and the loss function is L1-loss.

In order to save time and resources and improve the performance of the model, this paper selects ETH-Xgaze, a dataset with a large amount of data and a wide range of head pose and gaze, for pre-training. During pre-training, batch size, epoch, weight decay, and decay step are set to 64, 30, 0.5, and 10, respectively. the learning rate is adopted as Cyclical Learning Rate (CLR)\cite{b33}, base\_lr and max\_lr are 0.0001 and 0.0005, respectively, and the step size is set to 5. Since CLR allows the learning rate to fluctuate within a certain range, it helps to prevent the model from falling into a local optimum, and the change of the learning rate can help the model to be better generalized to new data, which improves the performance of the model. The optimizer uses AdamW \cite{b34}, betas =(0.88, 0.99). Pre-training begins with a linear learning rate warm-up, set to 3 epochs.

In the training phase, the batch size, epoch, and learning rate of Gaze360, MPIIFaceGaze, EyeDiap, and RT-Gene datasets are (32, 60, 0.0005), (64, 80, 0.0005), (12, 50, 0.0005), and (32, 50, 0.0005), respectively. The weight decay and decay step of the MPIIFaceGaze dataset are 0.5 and 60, respectively, and the rest of the datasets are not decayed.  The optimizer in all the above four datasets uses Adam \cite{b35} with betas = (0.88, 0.999). The training warms-up using a linear learning rate set for 5 epochs. Leave-one-out cross-validation, 4-fold cross-validation, and 3-fold cross-validation are used for the MPIIFaceGaze, EyeDiap, and RT-Gene datasets, respectively, since the MPIIFaceGaze, EyeDiap, and RT-Gene datasets do not have a strict separation between training and testing sets. In addition, due to the overall small size of the EyeDiap data, there is a risk of overfitting, so the training uses a linearly decaying dropout rate \cite{b42} in Stage 1, Stage 2, and Stage 3, which is 0.09, 0.06, and 0.03, respectively. This approach can effectively avoid overfitting of low-level features and lack of high-level semantics, thus improving the robustness and stability of model training.
\subsection{Comparison with classical methods}
To validate the performance of FGI-Net, comparisons are made with existing classical models and lightweight models. The comparison includes angle error, parameters, and FLOPs.
FGI-Net is first compared with existing classical models, and a rigorous experiment confirms that FGI-Net achieves significant improvements on various benchmarks. Compared to the classical method GazeCaps, FGI-Net improves the accuracy by 0.32°, 0.29°, and 0.90° on the three datasets MPIIFaceGaze, EyeDiap, and RT-Gene with 87.1\% fewer parameters and 79.1\% fewer FLOPs, respectively. Compared with the classical method GazeTR, FGI-Net achieves performance improvement on all datasets with 86.7\% fewer parameters and 79.2\% fewer FLOPs, especially on the RT-Gene dataset with an 8\% performance improvement, see Table 2 for detailed data.

\begin{table}[htbp]
\caption{Comparison with classical methods}
\begin{center}
\scalebox{1.1}{
\begin{tabular}{cccccc}
\hline
Methods   & G{\cite{b29}} & E{\cite{b31}} & R{\cite{b32}} & M{\cite{b30}} & Year \\ \hline
Dilated-Net{\cite{b19}} & 13.73°          & 6.19°           & 8.38°           & 4.42°                & 2018 \\
Gaze360{\cite{b29}}     & 11.04°          & 5.36°           & 7.06°           & 4.06°                & 2019 \\
CA-Ne{\cite{b37}}       & 11.20°          & 5.27°           & 8.27°           & 4.27°                & 2020 \\
LeViT-128S{\cite{b39}}  & N/A             & 5.62°           & N/A             & 5.33°                & 2021 \\
T2T-ViT-7{\cite{b40}}   & N/A             & 5.54°           & N/A             & 4.99°                & 2021 \\
GazeTR {\cite{b24}}     & 10.62°          & 5.17°           & 6.55°           & 4.00°                & 2022 \\
L2CS{\cite{b18}}        & 10.41°          & N/A             & N/A             & 3.92°                & 2022 \\
SwAT{\cite{b38}}        & 11.60°          & N/A             & N/A             & 5.00°                & 2022 \\
EFE{\cite{b5}}          & N/A             & N/A             & N/A             & 4.40°                & 2023 \\
GazeCaps{\cite{b36}}    & \textbf{10.04°}          & 5.44°           & 6.92°           & 4.06°                & 2023 \\ \hline
\textbf{FGI-Net(ours)}       & 10.50°          & \textbf{5.15°}           & \textbf{6.02°}           & \textbf{3.74°}                & 2024 \\ \hline
\end{tabular}
\label{tab2}
}
\end{center}
\end{table}
In order to compare FGI-Net with the lightweight models, this paper uses the MPIIFaceGaze and EyeDiap datasets to compare with the lightweight models Dilated-Net\cite{b19} ,LeViT-128S \cite{b39} and T2T-ViT-7 \cite{b40} in terms of angular error, parameters, and FLOPs dimensions. The experimental results are shown in Table 3, from which it can be seen that the angular error of FGI-Net is significantly smaller than the other lightweight models, and the parameters of FGI-Net are reduced by at least 61.5\%.

\begin{table}[htbp]
\caption{Comparison with Lightweight methods.}
\begin{center}
\begin{tabular}{ccccc}
\hline
Methods             & M{\cite{b30}}   & E{\cite{b31}}   & Paramaters(M) & FLOPs(G)      \\ \hline
Dilated-Net{\cite{b19}} & 4.42°          & 6.19°          & 3.92          & 3.15          \\
LeViT-128S{\cite{b39}}  & 5.33°          & 5.62°          & 7.00          & \textbf{0.28} \\
T2T-ViT-7{\cite{b40}}   & 4.99°          & 5.54°          & 4.00          & 0.98          \\ \hline
\textbf{FGI-Net(ours)}       & \textbf{3.74°} & \textbf{5.15°} & \textbf{1.51} & 0.38          \\ \hline
\end{tabular}
\label{tab3}
\end{center}
\end{table}

As can be seen from Tables 2 and 3, the FGI-Net network model has certain advantages compared with both classical and lightweight methods, mainly because the global information fusion module proposed in this paper can use the shifting window mechanism to interact with the global information in different periods of the network model, which makes FGI-Net not only utilize the local information but also fuse the global information, and the global information provides the a priori information related to the gaze estimation and can also be used to compensate for the eye gaze direction. In addition, The Res\_CBAM module enables FGI-Net to focus more on important features and better adapt to the structure and content of input images, thereby improving the overall performance of the model.

\begin{figure*}[htbp]
\centerline{\includegraphics[width=1\linewidth]{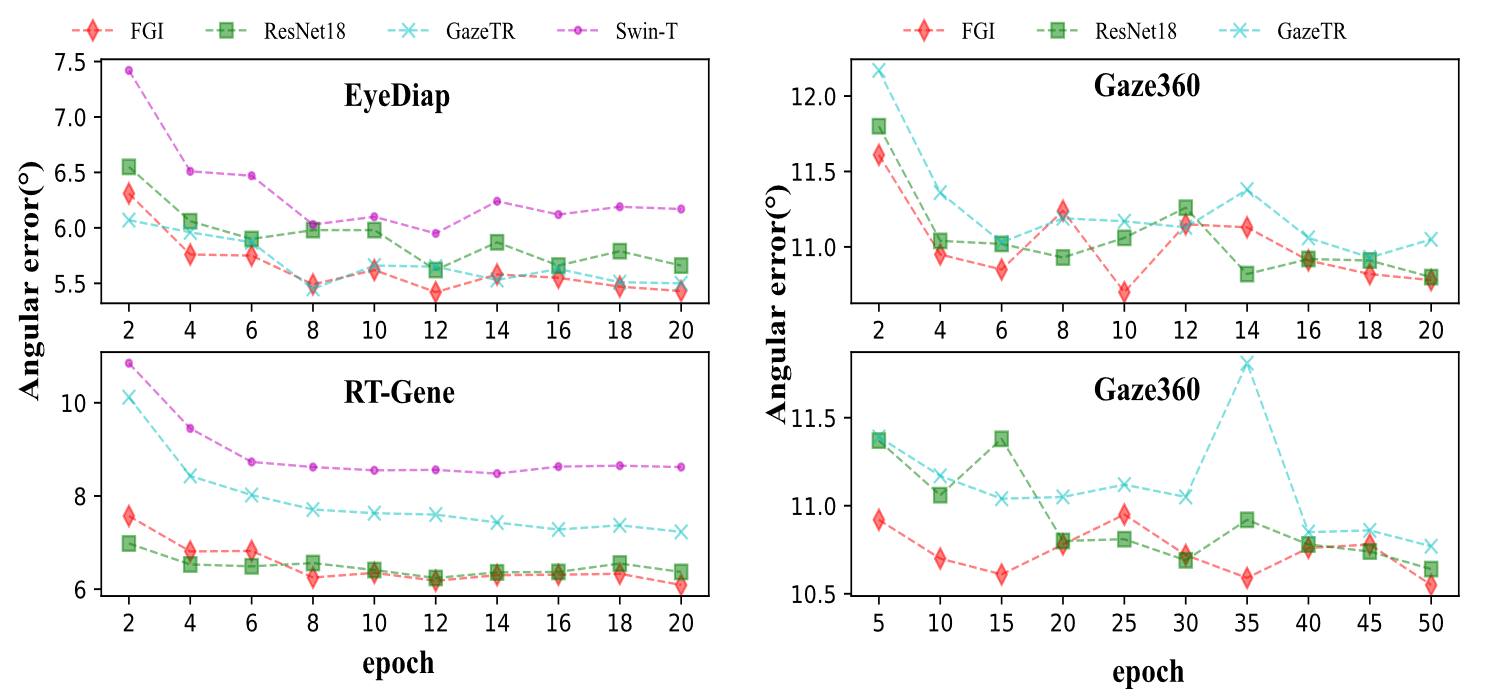}}
\caption{Model convergence trend}
\label{fig4}
\end{figure*}
\subsection{Comparison of model convergence speed}
In order to verify whether Global Information Fusion module can accelerate the convergence of the models, this paper selects the current mainstream models with different architectures (CNN, Transformer, and CNN+Transformer) and compares them with FGI-Net in terms of dimensions such as dataset size and testing methods. The detailed data of the models used in this section are shown in Table 4.
\begin{table*}[htbp]
\caption{Related parameters of different models.}
\begin{center}
\begin{tabular}{cccc}
\hline
Methods         & Backbone                        & Paramaters(M) & FLOPs(G)      \\ \hline
ResNet{\cite{b2}}   & ResNet18(CNN)                   & 11.18         & 1.82          \\
Swin{\cite{b6}}     & Swin-T(Transformer)             & 27.52         & 4.37          \\
GazeTR {\cite{b24}} & CNN+ Transformer                & 11.42         & 1.83          \\ \hline
\textbf{FGI-Net(ours)}   & Alternating CNN and Transformer & \textbf{1.51} & \textbf{0.38} \\ \hline
\end{tabular}
\label{tab4}
\end{center}
\end{table*}

In order to analyze the convergence trend of the model more intuitively, this paper selects the first 20 epochs for model training, and samples one point every 2 epochs to display it graphically(as shown in Fig.4). The experimental data in the first column of Fig.4 show that both the large dataset (RT-Gene) and the small dataset (EyeDiap) FGI-Net can converge faster and have smaller fluctuations when using one-fold cross-validation. Although ResNet18 exhibits convergence ability comparable to FGI-Net on large datasets, its convergence ability has decreased on small datasets. When the dataset is strictly divided into training and testing sets, the same tests are conducted on the Gaze360 dataset.  The second column of data in Fig.4 shows that FGI-Net does not exhibit significant convergence advantages in the first 20 epochs and the first 50 epochs, but FGI-Net fluctuates less, and from Table 4, it can be seen that the parameter quantity and FLOPs of FGI-Net are significantly reduced. Due to the significant performance difference of the Swin model on the Gaze360 dataset compared to other methods, relevant data is not shown. From the analysis of the Gaze360 dataset in Table 1, it can be seen that the head posture and gaze angle of the samples in the dataset are relatively large, and there are many subjects with a small number per capita. There is some correlation between the experimental results and the characteristics of the Gaze360 dataset.

\subsection{Ablation Study}
In order to better understand the impact of different parts on the performance of FGI-Net, ablation experiments are conducted, removing residual connections and Res\_CBAM from the backbone model. Experiments have shown that removing residual connections and Res\_CBAM lead to different levels of performance degradation, with the most severe degradation observed on the Gaze360 dataset. From Table 1, it can be seen that this dataset has significantly more participants than other datasets, but the per capita sample size is small, and the head posture and gaze angle are also much larger than other datasets. The performance degradation of RT-Gene and MPIIFaceGaze is relatively small. Analyzing the dataset, it can be seen that the per capita sample size of these two datasets is significantly larger than Gaze360, and the head posture and gaze angle are much smaller than Gaze360. Therefore, removing certain components will not cause significant performance degradation. The detailed data are shown in Table 5.
\begin{table}[htbp]
\caption{Ablation study results for the error angle.}
\begin{center}
\scalebox{1.1}{
\begin{tabular}{ccccc}
\hline
            & G{\cite{b29}} & E{\cite{b31}} & R{\cite{b32}} & M{\cite{b30}} \\ \hline
FGI-Net     & \textbf{10.50°}          & \textbf{5.15°}           & \textbf{6.02°}           & \textbf{3.74°}                \\
- Residual  & 10.87°          & 5.32°           & 6.06°           & 3.86°                \\
- Res\_CBAM & 10.93°          & 5.24°           & 6.16°           & 3.80°                \\ \hline
\multicolumn{5}{l}{Symbol ‘-’ indicates the following component is remove.}
\end{tabular}
\label{tab5}
}
\end{center}
\end{table}

\section{Conclusions and Analysis}
In this paper, we propose a gaze estimation model FGI-Net that fuses global information to maintain lightweight while comparing favorably with other classical models in terms of accuracy. The Global Information Fusion module in the FGI-Net model not only extracts features efficiently, but also expands the receptive field to link contextual information. FGI-Net is validated on four widely used datasets, and in the vast majority of cases, FGI-Net achieves the smallest angular error. In addition, FGI-Net is also compared with other lightweight models, and FGI-Net not only reduces the parameters by more than 61.5\% but also reduces the angular error significantly.

In future work, in order to further enhance the likelihood of the model's application in the real world, the CNN's advantage in feature extraction can be utilized to maintain the model's excellent performance, while taking advantage of the Transformer's large receptive field to reduce the stacking of convolutional and pooling layers thereby realizing the model's lightweight and reducing the loss of information. However, the output of both Transformer and CNN is scalar, and less information is preserved. In order to preserve more information, such as the relative positional relationship between features, Capsules Network \cite{b41} is further integrated into the model, taking advantage of the fact that both the input and output of Capsules Network are vectors, so that the model can utilize more effective information and achieve maximum performance.

\end{document}